\documentclass[sigconf]{acmart}

\copyrightyear{2026}
\acmYear{2026}
\setcopyright{cc}
\setcctype{by}
\acmConference[WWW '26]{Proceedings of the ACM Web Conference 2026}{April 13--17, 2026}{Dubai, United Arab Emirates}
\acmBooktitle{Proceedings of the ACM Web Conference 2026 (WWW '26), April 13--17, 2026, Dubai, United Arab Emirates}
\acmPrice{}
\acmDOI{10.1145/3774904.3792492}
\acmISBN{979-8-4007-2307-0/2026/04}

\usepackage{latexsym}
\usepackage{booktabs}
\usepackage{multirow}
\usepackage[ruled,vlined]{algorithm2e}
\usepackage{svg}
\usepackage{amsthm,amsmath}
\usepackage{mathrsfs}
\usepackage{tabularx}
\usepackage{enumitem}
\usepackage{makecell}
\usepackage{balance}

\begin{document}

\newcolumntype{L}[1]{>{\raggedright\arraybackslash}p{#1}}
\newcolumntype{C}[1]{>{\centering\arraybackslash}p{#1}}
\newcolumntype{R}[1]{>{\raggedleft\arraybackslash}p{#1}}

\title{Frequency-Corrupt Based Graph Self-Supervised Learning}



\author{Haojie Li}
\orcid{0009-0001-0863-9576}
\affiliation{%
  \institution{School of Data Science, Qingdao University of Science and Technology}
  \city{Qingdao}
  \country{China}}
\email{lihaojie@qust.edu.cn}

\author{Mengjiao Zhang}
\orcid{0009-0009-0748-5948}
\affiliation{%
  \institution{School of Information Science and Technology, Qingdao University of Science and Technology}
  \city{Qingdao}
  \country{China}
}
\email{zhang_mj@mails.qust.edu.cn}

\author{Guanfeng Liu}
\authornote{Corresponding authors.}
\orcid{0000-0001-8980-4950}
\affiliation{%
  \institution{School of Computing, Macquarie University}
  \city{Sydney}
  \country{Australia}
}
\email{guanfeng.liu@mq.edu.au}

\author{Qiang Hu}
\orcid{0000-0001-7642-5660}
\affiliation{%
  \institution{School of Information Science and Technology, Qingdao University of Science and Technology}
  \city{Qingdao}
  \country{China}
}
\email{huqiang@qust.edu.cn}

\author{Yan Wang}
\orcid{0000-0002-5344-1884}
\affiliation{%
  \institution{School of Computing, Macquarie University}
  \city{Sydney}
  \country{Australia}
}
\email{yan.wang@mq.edu.au}

\author{Junwei Du}
\authornotemark[1]
\orcid{0000-0002-2909-2565}
\affiliation{%
  \institution{School of Data Science, Qingdao University of Science and Technology}
  \city{Qingdao}
  \country{China}
}
\email{dujunwei@qust.edu.cn}

\renewcommand{\shortauthors}{Haojie Li et al.}

\begin{abstract}
Graph self-supervised learning (GSSL) alleviates the graph data labeling bottleneck without supervision, enabling wide application in domains like recommendation systems and social network analysis.
High-frequency signals are valuable in GSSL for capturing local structural preferences, thereby enriching graph representations and boosting model performance.
However, in practical applications, two critical problems hinder the efficient and robust use of these signals.
First, the locality of high-frequency signals limits their full utilization by the model.
Second, over-reliance on specific high-frequency signals will affect the model's generalization.
To address the above problems, we propose the \textbf{F}requency-\textbf{C}orrupt Based \textbf{G}raph \textbf{S}elf-\textbf{S}upervised \textbf{L}earning (FC-GSSL) algorithm.
Specifically, we generate corrupted graphs biased toward high-frequency signals by corrupting nodes and edges according to their low-frequency contributions.
These corrupted graphs are fed as input to an autoencoder, with low-frequency and general features serving as the supervision. 
This compels the model to effectively fuse high- and low-frequency signals, thereby integrating and utilizing more valuable high-frequency information.
Additionally, we design multiple sampling strategies and form diverse corrupted graphs based on the intersections and union between the results obtained from these strategies. 
By aligning the node representations from these views, the model can identify valuable frequency combinations, which helps reduce the negative impact of specific high-frequency components and improve generalization.
FC-GSSL optimizes the design of GSSL for web applications, significantly improving model performance on complex web-related graphs, such as social networks and citation networks.
This work makes a direct contribution to advancing the "Graph Algorithms and Modeling for the Web" research track.
Experimental results on 14 datasets across multiple tasks illustrate the superiority of the proposed approach. 

\end{abstract}




\maketitle




\section{Introduction}

GSSL
is broadly categorized into two paradigms: contrastive learning and generative learning \cite{wang2024uncovering}.
Contrastive learning methods capture structural patterns by maximizing mutual information between positive pairs in different views \cite{hassani2020contrastive,li2025intent}.
Meanwhile, generative learning methods learn structural details by reconstructing corrupted parts of the graph \cite{li2023s,liu2025graph}.
GSSL can learn expressive graph representations without manual labels to alleviate the annotation bottleneck, and is widely applied in recommendation systems \cite{li2025intent,li2025candidate}, social network analysis \cite{xu2024adaptive,singh2024social}, and drug discovery \cite{jiang2021could,bongini2021molecular}.


However, most GSSL methods \cite{wang2024rethinking, hou2022graphmae} still prefer low-frequency signals and overlook valuable high-frequency patterns, limiting the learning of expressive structural information \cite{liu2025graph}. To address this, recent studies focus on capturing high-frequency signals in graphs: for example, ChebyCF \cite{kim2025graph} and PolyCF \cite{qin2025polycf} apply graph spectral filtering to enhance high-frequency sensitivity, while GraphPAE \cite{liu2025graph} and HOPE-WavePE \cite{nguyen2024range} employ positional encodings to capture them effectively. Despite progress \cite{liu2025graph,qin2025polycf}, two critical problems hinder the efficient and robust use of these signals.

\noindent\textbf{Problem 1.}  
The locality of high-frequency signals limits their full utilization by the model.
Current deep learning models are mainly based on statistical inductive learning \cite{wang2024towards}. 
They extract stable patterns from the overall data distribution to obtain generalizable and transferable representations \cite{lei2023generalization}. 
However, in graph data, high-frequency components often reflect local preferences \cite{li2024matters,zou2025loha}. 
These signals are sparse, non-stationary, and highly sensitive to sampling and noise \cite{yang2024wavenet}. 
They lack statistical stability and fail to form consistent and representative patterns at the group level.
Therefore, a fundamental conflict exists between the locality of high-frequency signals and statistical inductive learning. 
In conventional training procedures, models tend to preserve smooth and globally consistent patterns, potentially misclassifying valuable high-frequency information as noise and suppressing it. 
This results in insufficient utilization of high-order signals, thereby further compromising the model's representational capacity.

\noindent\textbf{Problem 2.}  
Over-reliance on specific high-frequency signals will affect the model's generalization.
High-frequency signals in graph data, which often reflect local preferences and specific patterns \cite{zou2025loha}, contradict the model's inherent need for smoothness and thus reduce its generalization ability \cite{qin2025polycf}.
Although some studies employ ideal low-pass filters to enhance the model's ability to extract low-frequency components, these methods merely combine high- and low-frequency signals through simple aggregation \cite{cai2023lightgcl,shen2021powerful}.
They fail to effectively coordinate the relationships between signals of multiple frequencies, making it difficult to discern whether specific combinations of frequency signal patterns represent valid information or mere coincidental noise. 
Therefore, the model may still have excessive reliance on specific high-frequency signals. 
During training, the model may focus too much on these local details, leading to overfitting and thus affecting the model's generalization ability.

To address the above problems, we propose the \textbf{F}requency-\textbf{C}orrupt Based \textbf{G}raph \textbf{S}elf-\textbf{S}upervised \textbf{L}earning (FC-GSSL) algorithm.
Specifically, we analyze the contribution of low-frequency signals in nodes and edges to the graph structure, from a spectral perspective \cite{bo2023specformer,bo2023survey}.
Based on this, we corrupt nodes and edges according to their low-frequency contributions, creating corrupted graphs biased toward high-frequency signals. These graphs are fed into an autoencoder under a corruption-reconstruction paradigm, where supervision relies on both low-frequency and general features. This compels the model to learn how to transform high-frequency signals into lower-frequency representations and to more effectively explore their collaborative relationships.
Consequently, the model shifts its focus from merely extracting specific frequency signals to effectively fusing multi-frequency signals. This mechanism enhances the integration of valuable high-frequency signals, mitigating their underutilization. Additionally, we design multiple sampling strategies and form diverse corrupted graphs based on intersections and unions among their outcomes.
Through aligning node representations from corrupted graphs, the model improves its ability to distinguish meaningful multi-frequency signals from random noise. This reduces the negative influence of specific high-frequency components, enhancing the model’s robustness and generalization.
To sum up, the contributions of our work are summarized as follows:
\begin{itemize}[leftmargin=10pt]
\item
We construct high-frequency-biased graphs as autoencoder inputs and set low-frequency and general features as reconstruction targets, compelling the model to fuse multi-frequency signals and better utilize valuable high-frequency information.

\item
We generate diverse corrupted graphs via multiple sampling strategies and align their node representations, enabling the model to identify valuable frequency combinations, mitigate specific high-frequency effects, and enhance generalization.

\item
We compare FC-GSSL with state-of-the-art baseline methods across multiple tasks, including node classification, graph property prediction, and transfer learning. Results show FC-GSSL consistently outperforms the baseline methods on 14 graphs, with higher accuracy and better generalization in most tasks.

\end{itemize}

\section{Related Work}

\subsection{Graph Self-supervised Learning }

\noindent\textbf{Graph Contrastive Learning.}
Graph contrastive learning enhances model robustness through the alignment of positive sample pairs from different views. This process captures semantic invariance and reduces irrelevant information in node representations \cite{hassani2020contrastive,li2025intent}. 
For examples:
SGL \cite{wu2021self} and AutoGCL \cite{yin2022autogcl} achieve semantic-invariant learning by employing data augmentation techniques, such as node dropping and edge perturbation, and aligning representations of the same node across different views.
MSSGCL \cite{liu2024multi} and GraphCL \cite{you2020graph} leverage subgraph sampling and multi-scale contrastive learning to achieve fine-grained representation of node semantics.
NCL \cite{lin2022improving}, AdaCML \cite{zhu2023adamcl}, and IPCCF \cite{li2025intent} leverage contrastive learning to align nodes with their neighbors in user-item graphs, generating better representations for recommendation.

\noindent\textbf{Graph Generative Learning.}
Graph generative learning methods, such as graph autoencoders, require the model to reconstruct corrupted parts by leveraging available context, which facilitates the learning of intricate structural details within the graph \cite{li2023s,liu2025graph}.
For examples:
VGAE \cite{kipf2016variational} and GATE \cite{salehi2019graph} achieve self-supervised learning by corrupting and reconstructing links and features, respectively.
GraphMAE \cite{hou2022graphmae} and GraphMAE2 \cite{hou2023graphmae2}, as masked graph autoencoders, address the issue of excessive reliance on neighboring information in traditional graph autoencoders through efficient feature masking and reconstruction, thereby improving the performance of generative models on various downstream tasks.
HGMAE \cite{tian2023heterogeneous} and GraMI \cite{zhao2025variational}, using node attribute corruption and reconstruction, effectively encode complex heterogeneous graphs.


While above methods have shown promise, they mainly focus on low-frequency signals and neglect local preference patterns, 
which limits the ability to capture expressive structural information \cite{liu2025graph}.

\subsection{Usage of High-Frequency Signals}

\noindent\textbf{Graph Spectral Filtering.}
Graph spectral filtering employs effective spectral filters to extract high-frequency signals, thereby capturing local information in graphs \cite{qin2025polycf, liu2022revisiting}.
For examples:
DSF \cite{guo2023graph} introduces a diversified spectral filtering framework that learns node-specific filter weights, thereby leveraging distinct local structural features.
Sp$^2$GCL \cite{bo2023graph} fuses spatial and spectral graph views through consistency capture between GNN-encoded spatial information and robust spectral features.
BernNet \cite{he2021bernnet}, JacobiConv \cite{wang2022powerful}, and ChebyCF \cite{kim2025graph} leverage high-order polynomial graph filters to express arbitrarily smooth functions, thereby achieving effective extraction of diverse graph frequency signals.
GF-CF \cite{shen2021powerful} and PGSP \cite{liu2023personalized} enhance the ability to capture frequency signals by integrating graph filter properties into collaborative filtering.

\noindent\textbf{Positional Encodings.}
Infusing position information associated with high-frequency signals during graph encoding can enhance the model's capacity to capture local graph details, thus boosting its representational power \cite{liu2025graph}.
For examples: 
Some studies \cite{dwivedi2023benchmarking,dwivedi2021graph}use the eigenvectors of the graph Laplacian matrix as initial node positional encodings to represent relative node relationships in the spectral domain.
RFP \cite{eliasof2023graph} enhances model expressiveness by using random feature propagation instead of directly computing Laplacian eigenvectors, which helps preserve high-frequency components.
SPE \cite{huang2023stability} perturbs the Laplacian matrix to generate robust positional encodings for enhanced high-frequency signal extraction.
PEG \cite{wangequivariant} uses dual channels for updates of independent nodes and positional characteristics, with bidirectional constraints to resolve ambiguities of the sign and basis to improve high-frequency use.


Despite their success with high-frequency signals, the above methods are limited by signal locality and an over-reliance on specific signals, which hinders their overall effectiveness.


\section{Preliminaries}
In this section, we first define the relevant notations and basic concepts to facilitate the elaboration of our subsequent method.

\noindent\textbf{Problem Definition.}
Given a graph $\mathcal{G}=\{ \mathcal{V}, \mathcal{E}, \textbf{X} \},$
where $\mathcal{V}$ denotes the node set with $ \mathcal{V} = \{ v_i \}_{i=1}^N$, 
$\mathcal{E} \subseteq \mathcal{V} \times \mathcal{V}$
denotes the edge set
with $e_{ij} \in \mathcal{E}$ connecting node $v_i$ and node $v_j$
, and $\textbf{X} \in \mathbb{R}^{N \times d}$ is the node feature matrix with $d$-dimensional feature vectors.
The graph topology is represented by an adjacency matrix $\textbf{A} \in \{0,1\}^{N \times N}$, where 
$A_{ij}=1$ if an edge exists between nodes 
$v_i$ and $v_j$, and 0 otherwise.
The objective of our work is to learn an unsupervised graph encoder 
$f_{enc}(\cdot)$ that maps the input graph to meaningful node representations 
$\textbf{H} = f_{enc}(\mathcal{G}) \in \mathbb{R}^{N \times d_h}$ without using labeled data.

\noindent\textbf{Laplacian Matrix, Eigenvalues and Eigenvectors.}
The normalized graph Laplacian is defined as 
$\textbf{L} = \textbf{I}_N - \textbf{D}^{-1/2} \textbf{A}  \textbf{D}^{-1/2}$, where 
$\textbf{I}_N$ is the identity matrix and 
$\textbf{D}$ is the diagonal degree matrix constructed from the graph's adjacency matrix 
$\textbf{A}$, with each diagonal entry 
$D_{ii}$ equal to the degree of node 
$v_i$.
The Laplacian matrix 
$\textbf{L}$ admits an eigen-decomposition 
$\textbf{L} = \textbf{U} \Lambda \textbf{U}^T$, where 
$\textbf{U} = [\textbf{u}_1, \textbf{u}_2, \dots, \textbf{u}_N ]$ is an orthogonal matrix whose columns are the eigenvectors of $\textbf{L}$, and 
$\Lambda = diag(\lambda_1,\lambda_2, \dots, \lambda_N)$ is the diagonal matrix of the corresponding eigenvalues. 
These satisfy the relation 
$\textbf{L}\textbf{u}_i = \lambda_i \textbf{u}_i$ for each $i$.

\section{Methodology}


\begin{figure*}[h]
\setlength{\abovecaptionskip}{0pt}
\centering
\includegraphics[scale=0.40]{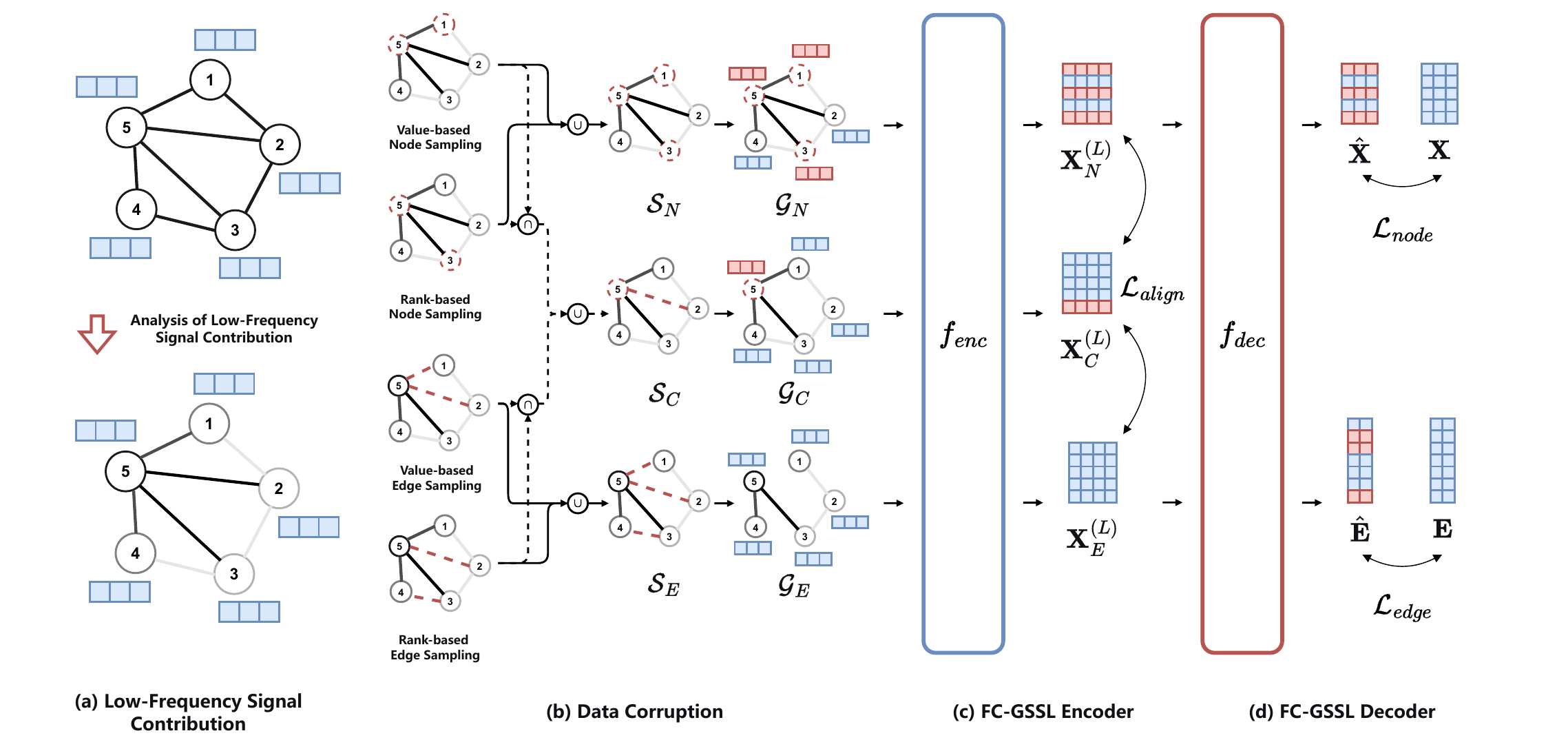}
\caption{
An overview of FC-GSSL
}
\label{fig:label}
\end{figure*}

\subsection{Overview}
The overall framework of FC-GSSL, as shown in Fig. 1,
consists of four modules:
\begin{itemize}[leftmargin=10pt]

\item
\textbf{Low-Frequency Signal Contribution Module}:
This module analyzes low-frequency signal contributions via the Laplacian matrix, generating high-frequency biased corrupted graphs.

\item \textbf{Data Corruption Module}:
This module develops and combines sampling strategies from low-frequency signal contributions for generating diverse high-frequency biased corrupted graphs.

\item \textbf{FC-GSSL Encoder Module}:
This module processes corrupted graphs through the encoder and aligns the resulting multi-view node representations to mitigate over-reliance on specific frequency signals and enhance generalization.

\item \textbf{FC-GSSL Decoder Module}:
This module decodes node representations and reconstructs corrupted data from both node features and edge structure, enabling the model to learn collaborative multi-frequency signal relationships for effective fusion.

\end{itemize}

\subsection{Low-Frequency Signal Contribution}


To enhance the model's use of high-frequency signals, we analyze low-frequency contributions in nodes and edges to generate corrupted graphs biased toward high-frequency information.

\noindent\textbf{Analysis of Low-Frequency Signal Contribution in Edges.}
To analyze the contribution of low-frequency signals to edges, we propose a novel metric named the area under the curve of low-frequency signal contribution in edges  ($C_E$), which is inspired by the AUC concept \cite{li2024area}.
This metric aggregates signal contributions across all frequencies using the following formula:
\begin{equation}
\small
C_{E,ij} = \frac{1}{K} \sum_{m=1}^{K} \frac{ 
\sum_{n=1}^m \lvert \textbf{u}_{ni} \times \lambda_n \times \textbf{u}_{nj} \rvert }{
\sum_{n=1}^K \lvert \textbf{u}_{ni} \times \lambda_n \times \textbf{u}_{nj} \rvert
},
\end{equation}
where $C_{E,ij}$ represents the contribution value of low-frequency signals in edge $e_{ij}$ to the graph structure, $K$ denotes the number of top-frequency components we are concerned with, $\textbf{u}_{ni}$ denotes the $i$-th element in eigenvector $\textbf{u}_{n}$, and $\lambda_n$ represents the $n$-th eigenvalue.

\noindent\textbf{Analysis of Low-Frequency Signal Contribution in Nodes.}
After obtaining low-frequency signal contributions in edges, we aggregate them around each node to compute the low-frequency contribution in nodes $C_N$, as follows:
\begin{equation}
\small
C_{N,i} = \frac{1}{|\mathcal{N}_{(i)}|}
\sum_{j \in \mathcal{N}_{(i)}} C_{E,ij}
\end{equation}
where $C_{N,i}$ represents the contribution value of low-frequency signals in node $v_{i}$ to the graph structure, 
and $\mathcal{N}_{(i)}$ represents the set of neighboring indices of node $v_i$.

\subsection{Data Corruption}

In this paper, we corrupt the original graph data based on low-frequency signal contributions to obtain corrupted graphs biased towards high-frequency signals.
The process consists of two steps: corrupt item selection and corrupt target generation.

\subsubsection{Corrupt Item Selection}

In this section, we develop multiple sampling strategies based on the value and rank of low-frequency signal contributions. 
The final corruption targets are determined by combining the results from these strategies.

\noindent\textbf{Value-based Sampling Strategy.}
In this strategy, we perform the sampling operation based on the contribution of low-frequency signals in nodes or edges to the graph structure, such that nodes or edges with higher contribution values have a higher probability of being sampled, which is defined as follows:
\begin{equation}
\small
\mathcal{P} \sim Multinomial(C,T).
\end{equation}
In this process, $T$ items are drawn without replacement from the entire set, with the selection probability for each item proportional to its contribution value $C$. 
This is equivalent to sampling indices from a multinomial distribution. 
These sampled indices constitute the sampled item set $\mathcal{P}$.
In practice, the value of $T$ is determined by the sampling rate $r$.

\noindent\textbf{Rank-based Sampling Strategy.}
To avoid over-reliance on the specific calculation method and numerical values of signal contributions, we designed a rank-based sampling strategy alongside the value-based approach.
Specifically, we sort the contribution values by their magnitude and replace the original values with their corresponding rank indices. 
The specific formula is as follows: 
\begin{equation}
\small
\mathcal{Q} \sim Multinomial(RI(C),T),
\end{equation}
where 
$RI(\cdot)$ is defined as the ranking and indexing operation, and 
$\mathcal{Q}$ is the sampled item set derived from the rank-based sampling strategy.

\noindent\textbf{Corrupted Item Selection Strategy.}
To mitigate the model’s over-reliance on specific high-frequency signals and the bias induced by specific sampling strategies, we draw on the Jaccard Similarity Coefficient \cite{park2023collaborative}, which asserts that the similarity between two sets increases with the ratio of the size of their intersection to the size of their union.
Therefore, after generating sampled item sets based on value-based and rank-based sampling strategies, we aim for the data corruption caused by the intersection and union of these sets to have approximately the same impact on the model representations. 
This enables the model to integrate high-frequency signals while mitigating its over-reliance on specific high-frequency signals.
Based on the above analysis, we construct the sampled item sets 
$\mathcal{S}_N= \mathcal{P}_N \cup \mathcal{Q}_N$, 
$\mathcal{S}_{\widetilde{N}}= \mathcal{P}_N \cap \mathcal{Q}_N$, 
$\mathcal{S}_E= \mathcal{P}_E \cup \mathcal{Q}_E$, and 
$\mathcal{S}_{\widetilde{E}}= \mathcal{P}_E \cap \mathcal{Q}_E$ with different sampling strategies, considering the contributions of low-frequency signals from nodes ($C_N$) and edges ($C_E$) to the graph structure.
Specifically,
$\mathcal{P}_N$ and 
$\mathcal{Q}_N$ represent the sampled item sets derived from $C_N$ 
using the value-based and rank-based sampling strategies, respectively.
Correspondingly, the sampling rates for nodes and edges are $r_N$ and $r_E$, respectively.
Furthermore, we design a comprehensive sampled item set $\mathcal{S}_C= \mathcal{S}_{\widetilde{N}} \cup \mathcal{S}_{\widetilde{E}}$.
By ensuring that the data corruption effects on the model are similar between sampled item sets $\mathcal{S}_N$ and $\mathcal{S}_C$ and between sampled item sets $\mathcal{S}_E$ and $\mathcal{S}_C$, this method reduces the model's over-reliance on specific high-frequency signals while enhancing its stability. 

\subsubsection{Corruption Target Generation}
After obtaining the sampled item sets, we perform data corruption operations based on the types of items in the sets to generate the corrupted graphs.

For node-type items, we perform feature masking.
We follow GraphMAE \cite{hou2022graphmae} by replacing the features of the sampled nodes $\widetilde{\mathcal{V}} = \mathcal{S} \cap \mathcal{V}$ with a learnable vector,
where $\mathcal{S}$ represents the sampled item set on which we operate and $\mathcal{V}$ represents the original node set.
The corrupted feature matrix $\widetilde{\textbf{X}}$ is defined as follows:
\begin{equation}
\small
\widetilde{\textbf{X}}_i = 
\begin{cases}
\textbf{x}_{[M]}, & \text {if $v_i \in \widetilde{\mathcal{V}}$} \\
\textbf{X}_i, & \text {if $v_i \notin \widetilde{\mathcal{V}}$} 
\end{cases}
\end{equation}
where $\textbf{X}_i \in \mathbb{R}^{d}$ denotes the original feature of node $v_i$, and $\textbf{x}_{[M]} \in \mathbb{R}^{d}$
denotes a learnable vector.

For edge-type items, we perform edge dropping.
We obtain the corrupted edge set $\widetilde{\mathcal{E}}$ by the formula $\widetilde{\mathcal{E}} = \mathcal{E} \backslash \mathcal{S}$, where $\mathcal{S}$ represents the sampled item set on which we operate, and $\mathcal{E}$ represents the original edge set.

After introducing the specific data corruption methods, we corrupted the sampled item sets 
$\mathcal{S}_N$, 
$\mathcal{S}_E$, and 
$\mathcal{S}_C$, 
yielding the corresponding corrupted graphs 
$\mathcal{G}_N$, 
$\mathcal{G}_E$, and 
$\mathcal{G}_C$.

\subsection{FC-GSSL Encoder}
In this section, we detail the encoding and representation alignment process of FC-GSSL.
In the encoding phase, we follow the GraphPAE \cite{liu2025graph} to integrate node features and pairwise distances via an attention mechanism, allowing effective fusion of multi-frequency signals, which can be formulated as:
\begin{equation}
\small
\textbf{X}^{(l+1)}, \textbf{P}^{(l+1)} = f_{enc}^{(l+1)}
\left(  
\textbf{X}^{(l)}, \textbf{P}^{(l)}
\right),
\end{equation}
where $l$ denotes the encoder layer, ranging from 0 to $L$,
and $f_{enc}^{(l+1)}$ denotes the process where the vector representation of layer $l$generates that of layer $l+1$ via message passing.
$\textbf{X}^{(l)} \in \mathbb{R}^{d_h}$ is the node feature matrix at the $l$-th layer, where its initial form 
$\textbf{X}^{(0)} \in \mathbb{R}^{d_h}$ is derived by mapping the original node feature matrix 
$\textbf{X} $ to a
$d_h$-dimensional space using a multi-layer perceptron.
$\textbf{P}^{(l)} \in \mathbb{R}^{d_h}$ denotes the node relative positional representation at the $l$-th layer. 
Its initial form $\textbf{P}^{(0)} \in \mathbb{R}^{d_h}$ is derived from the node relative position matrix $\textbf{P} \in \mathbb{R}^{N \times N}$, which is defined as:
\begin{equation}
\small
\textbf{P}_{ij} = 
\begin{cases}
\Vert \textbf{U}_i - \textbf{U}_j \Vert_2, & \text {if $\textbf{A}_{ij}=1$}, \\
0, & \text {otherwise}. 
\end{cases}
\end{equation}
We transform matrix $\textbf{P}$ into the representation $\textbf{P}^{(0)}$ through a series of Gaussian RBF kernels, using the operation
\begin{equation}
\small
\textbf{P}^{(0)}_{ij} = \text{MLP} \left( \left[ G(\textbf{P}_{ij};\mu_1,\sigma), \dots, G(\textbf{P}_{ij};\mu_d,\sigma)  \right]  \right),
\end{equation}
where MLP is another multilayer perceptron and
$G(\textbf{P}_{ij};\mu_k,\sigma) = \exp \left( 
-\left( \textbf{P}_{ij} -\mu_k  \right)^2 / 2 \sigma^2
\right)$ 
is the $k$-th Gaussian basis function, parameterized by a mean of $\mu_k$ and a standard deviation of $\sigma$.

\noindent\textbf{Message Passing Process.}
For the feature objective, we define the feature message passing path as 
\begin{equation}
\small
\begin{aligned}
& \textbf{X}_i^{(l+1)} = \sum_{j \in \mathcal{N}_{(i)}} \left( \alpha^{(l)}_{ij} + \textbf{P}^{(l)}_{ij}  \right) \odot \textbf{X}_j^{(l)},  \\
& \alpha^{(l)}_{ij} = f_{att} \left( \textbf{X}_i^{(l)},  \, \textbf{X}_j^{(l)}\right), \alpha^{(l)}_{ij} \in \mathbb{R}^{d_h},
\end{aligned}
\end{equation}
where $f_{att}$ is the attention function that calculates weights between neighboring nodes, and $\odot$ represents an element-wise multiplication.
The specific implementation of this attention mechanism varies with the underlying graph encoder architecture. When using a GAT \cite{velivckovic2017graph} as encoder, the parameter $d_h$ signifies the number of attention heads, and the corresponding function is defined as
\begin{equation}
\small
\alpha_{ij}^{(l)} = \text{LeakyReLU}
\left( 
\textbf{w}^T \left[
\textbf{W}^{(l)} \textbf{X}_i^{(l)}
\Vert
\textbf{W}^{(l)} \textbf{X}_j^{(l)}
\right]
\right),
\;
\textbf{w} \in \mathbb{R}^{2d_h}
.
\end{equation}
In contrast, when the GatedGCN architecture is used, $d_h$ is assigned a value equal to the dimensionality of the node embeddings, and the functional relationship is defined as
\begin{equation}
\small
\alpha_{ij}^{(l)} = \text{Sigmoid}
\left(
\textbf{W}^{(l)} \textbf{X}_i^{(l)}
+
\textbf{W}^{(l)} \textbf{X}_j^{(l)}
\right)
.
\end{equation}
This path takes into account the weights of both features and positions during the message passing process, allowing for a more refined fusion of multi-dimensional frequency signals.

For the position objective, we define the position message passing path as
\begin{equation}
\small
\textbf{P}^{(l+1)}_{ij} = \textbf{P}^{(l)}_{ij} + \alpha^{(l)}_{ij}.
\end{equation}
We use attention weights to continuously optimize positional representations by integrating more feature information. 
This process allows the positional representations to better approximate the actual node positions, thereby improving the accuracy of the model representations.

\noindent\textbf{Generation and Alignment of Node Representations.}
After detailing the implementation of FC-GSSL, we use $\mathcal{G}_N$, 
$\mathcal{G}_E$, and 
$\mathcal{G}_C$ as inputs to the encoder to obtain the $L$-layer vector representations 
$\textbf{X}^{(L)}_N$, $\textbf{X}^{(L)}_E$, and $\textbf{X}^{(L)}_C$ of the graph nodes, which serve as the final node embeddings, respectively.
To align node representations between different corrupted graphs, a contrastive objective based on InfoNCE loss \cite{chen2020simple} is employed. 
This objective functions by maximizing the similarity between representations of the same node from different corrupted graphs while minimizing the similarity between those of dissimilar nodes, formulated as:
\begin{equation}
\small
\mathcal{I}(\textbf{X}', \textbf{X}'') =
\frac{1}{N}
\sum_{i=0}^N
-\log
\frac{
\exp(s(\textbf{x}'_{i}, \textbf{x}''_{i}) / \tau)
}{
\sum_{i'=0}^N \exp(s(\textbf{x}'_{i}, \textbf{x}''_{i'}) / \tau)
},
\end{equation}
where $\textbf{x}'_{i}$ denotes the representation of the $i$-th node in the feature matrix $\textbf{X}'$, $s(\cdot, \cdot)$ indicates the similarity function between nodes, and $\tau$ represents the temperature parameter.
By aligning the node vector representations $\textbf{X}^{(L)}_N$ with $\textbf{X}^{(L)}_C$, and $\textbf{X}^{(L)}_E$ with $\textbf{X}^{(L)}_C$, we force the model to avoid over-reliance on specific high-frequency signals. 
The loss function is defined as follows:
\begin{equation}
\small
\mathcal{L}_{align} = \mathcal{I}(\textbf{X}^{(L)}_N,\textbf{X}^{(L)}_C)
+ \mathcal{I}(\textbf{X}^{(L)}_E,\textbf{X}^{(L)}_C).
\end{equation}

\subsection{FC-GSSL Decoder}
Depending on the type of corruption operation, we reconstruct the corrupted information from two sources: the feature of the corrupted nodes and the structure of the corrupted edges.

\noindent\textbf{Reconstructing the Features of Corrupted Nodes.} 
For the corrupted graph $\mathcal{G}_N$, we recover the corrupted information by reconstructing the node features, where $\mathcal{S}_N$ denotes the set of corrupted nodes, $\textbf{X}$ denotes the original node feature matrix, and $\textbf{X}^{(L)}_N$ denotes for the encoded node representations.
Employing a feature decoder, we map the node representations back to the original feature space. The process is as follows:
\begin{equation}
\small
\hat{\textbf{X}} = f_{dec}^N \left( \textbf{X}^{(L)}_N  \right),
\end{equation}
where $\hat{\textbf{X}}$ represents the reconstructed feature matrix.
Following GraphMAE and GraphPAE, we adopt the scaled cosine error (SCE) as the feature reconstruction loss
\begin{equation}
\small
\mathcal{L}_{node} = 
\frac{1}{\lvert  \mathcal{S}_N \rvert}
\sum_{v_i \in \mathcal{S}_N}
\left(
1 - \frac{
\textbf{X}_i^T \hat{\textbf{X}}_i
}{
\Vert \textbf{X}_i \Vert \cdot \Vert \hat{\textbf{X}}_i \Vert
}
\right)^{\gamma},
\end{equation}
where $\gamma \geq 1$ is a hyperparameter.

\noindent\textbf{Reconstructing the Structure of Corrupted Edges.}
For the corrupted graph $\mathcal{G}_E$, we recover the corrupted information by reconstructing the edge feature representations associated with its edges, where $\mathcal{S}_E$ denotes the set of corrupted edges, and $\textbf{X}^{(L)}_E$ represents the encoded node representations.
For an edge $e_{ij}$, its original edge feature can be denoted as $\textbf{E}_{ij} = \textbf{U}_i \odot \textbf{U}_j$, and its representation derived from the learned node embeddings can be denoted as $\Bar{\textbf{E}}_{ij} = \textbf{X}^{(L)}_{E,i} \odot \textbf{X}^{(L)}_{E,j}$.
We also construct a decoder to achieve the transformation of edge features, with the process as follows:
\begin{equation}
\small
\hat{\textbf{E}} = f_{dec}^E \left( \Bar{\textbf{E}}  \right),
\end{equation}
where $\hat{\textbf{E}}$ and $\textbf{E}$ share the same feature dimension.
Then, the edge structure reconstruction loss is defined as follows:
\begin{equation}
\small
\mathcal{L}_{edge} = 
\frac{1}{\lvert  \mathcal{S}_E \rvert}
\sum_{e_{ij} \in \mathcal{S}_E}
\left(
1 - \frac{
\textbf{E}_{ij}^T \hat{\textbf{E}}_{ij}
}{
\Vert \textbf{E}_{ij} \Vert \cdot \Vert \hat{\textbf{E}}_{ij} \Vert
}
\right)^{\gamma}.
\end{equation}

Having described all the components of FC-GSSL, its overall loss can be formulated as a weighted combination of multiple losses
\begin{equation}
\small
\mathcal{L} = 
\mathcal{L}_{node} 
+ \alpha \mathcal{L}_{edge}
+ \beta \mathcal{L}_{align},
\end{equation}
where $\alpha$ and $\beta$ are hyperparameters.
Detailed training procedure for FC-GSSL is provided in \textbf{Appendix \ref{fcgssl}}.

\subsection{Model Analysis}

\subsubsection{Analysis of the Design Rationale of FC-GSSL}

GSSL integrates two paradigms: contrastive learning, which aligns positive samples across views to enhance robustness \cite{hassani2020contrastive,li2025intent}, and generative learning, which reconstructs corrupted structures to capture graph details \cite{li2023s,liu2025graph}. 
Based on these premises, we propose the FC-GSSL model with the following design:

\begin{itemize}[leftmargin=10pt]

\item
We introduce a corrupted graph with high-frequency bias into an autoencoder, supervising the decoder with low-frequency and general features. This design forces the model to learn structural details and transform high-frequency signals into low-frequency representations, thereby improving multi-frequency fusion, enhancing cross-frequency collaborative understanding, and strengthening high-frequency information utilization.

\item
We generate multiple corrupted graphs through varied sampling strategies, using intersections and unions of their results. By aligning node representations across these graphs, we encourage the model to learn invariant semantic information, thereby enabling effective discrimination between high-frequency signals and noise. This reduces overreliance on specific frequency signals and improves the model’s generalization capability.

\end{itemize}

\subsubsection{Comparison with Existing Methods}

In this section, we compare FC-GSSL with existing methods that utilize high-frequency signals to evaluate the effectiveness and advantages of our model in leveraging such signals.

\begin{itemize}[leftmargin=10pt]

\item \noindent\textbf{Comparison with Existing Spectral Filtering Methods.}
Compared with methods PolyCF \cite{qin2025polycf} and Sp$^2$GCL \cite{bo2023graph}, our focus shifts from designing effective spectral filters to extract high-frequency signals, to developing a mechanism that forces the model to learn collaborative relationships among different frequency signals. By facilitating effective aggregation, this mechanism mitigates the conflict between the locality of high-frequency signals and statistical inductive learning, thereby enhancing their utilization.

\item \noindent\textbf{Comparison with Existing Positional Encoding Methods.}
Compared with methods GraphPAE \cite{liu2025graph} and HOPE-WavePE \cite{nguyen2024range}, we construct corrupted graphs biased toward high-frequency components and guide the model in learning the transformation to low-frequency and general features, thus improving positional encoding efficiency and promoting frequency-domain feature fusion.
In addition, contrastive learning is introduced to reduce reliance on specific frequencies and enhance generalization.

\end{itemize}

\noindent The time complexity analysis is provided in \textbf{Appendix \ref{tca}}.


\section{Experiments}


In this section, we conduct extensive experiments on node classification, graph prediction, and transfer learning to evaluate FC-GSSL. Dataset statistics and key model parameters are provided in \textbf{Appendices \ref{esd}} and \textbf{\ref{ehp}}, respectively. Detailed hyperparameter configurations for replication are available in the source code at https://github.com/rookitkitlee/FC-GSSL.


\begin{table*}
\setlength{\abovecaptionskip}{5pt}
\setlength{\belowcaptionskip}{-5pt}
\caption{Node classification results of different graph self-supervised learning, mean accuracy (\%) $\pm$ standard deviation. Bold
indicates the best performance and underline means the runner-up.}

\scriptsize
\centering
\resizebox{16cm}{!}{
\begin{tabular}{C{1.5cm}C{1.6cm}C{1.6cm}C{1.6cm}C{1.6cm}C{1.6cm}C{1.6cm}}
\toprule
\multirow{2}*{\textbf{Dataset}}  & \multicolumn{4}{c}{Small Graphs}    & \multicolumn{2}{c}{Large Graphs}      \\
\cmidrule(lr){2-5}   \cmidrule(lr){6-7}

& BlogCatalog & Chameleon & Squirrel  & Actor &  arXiv-year  & Penn94  \\
\midrule

DGI & 72.07$\pm$0.16 & 43.83$\pm$0.14 & 34.56$\pm$0.10 & 27.98$\pm$0.09 & - & -  \\
BGRL & 79.74$\pm$0.46 & 61.24$\pm$1.07 & 43.24$\pm$0.52 & 26.61$\pm$0.57 & 41.43$\pm$0.04 & 63.31$\pm$0.491  \\
MVGRL & 63.24$\pm$0.94 & 73.19$\pm$0.42 & 60.09$\pm$0.44 & 34.64$\pm$0.20 & -  & -  \\
CCA-SSG & 74.00$\pm$0.28 & 75.00$\pm$0.75 & 61.58$\pm$1.98 & 27.79$\pm$0.58 & 40.78$\pm$0.01 & 62.63$\pm$0.20 \\
Sp$^2$GCL & 72.73$\pm$0.46 & 78.88$\pm$1.04 & 62.61$\pm$0.87 & 34.70$\pm$0.92 & 39.09$\pm$0.02 & 68.80$\pm$0.45 \\
\midrule

VGAE & 60.47$\pm$1.84 & 62.32$\pm$1.90 & 42.50$\pm$1.35 & 31.57$\pm$0.75 & 36.39$\pm$0.21 & 55.31$\pm$0.28 \\
GraphMAE & 79.90$\pm$1.13 & 79.50$\pm$0.57 & 61.13$\pm$0.60 & 32.15$\pm$1.33 & 40.30$\pm$0.04 & 67.97$\pm$0.21 \\
GraphMAE2 & 77.34$\pm$0.12 & 79.13$\pm$0.19 & 70.27$\pm$0.88 & 34.48$\pm$0.26 & 38.97$\pm$0.03 & 67.86$\pm$0.42 \\
MaskGAE & 73.10$\pm$0.08 & 74.50$\pm$0.87 & 68.53$\pm$0.44 & 33.44$\pm$0.34 & 40.59$\pm$0.04 & 63.84$\pm$0.03 \\
S2GAE & 75.76$\pm$0.43 & 60.24$\pm$0.37 & 68.60$\pm$0.56 & 26.17$\pm$0.38 & 40.32$\pm$0.12 & 70.24$\pm$0.09 \\
AUG-MAE & 82.03$\pm$0.69 & 70.10$\pm$1.88 & 62.57$\pm$0.67 & 33.42$\pm$0.38 & 37.10$\pm$0.13 & 69.90$\pm$0.43 \\
GraphPAE & \underline{85.76$\pm$1.22} & \underline{80.51$\pm$1.25} & \underline{72.05$\pm$1.40} & \underline{38.55$\pm$1.35} & \underline{41.85$\pm$0.04}  & \underline{71.79$\pm$0.37} \\
\midrule


\textbf{FC-GSSL(Ours)} 
& \textbf{88.70}$\pm$\textbf{1.10} 
& \textbf{82.31}$\pm$\textbf{0.51} 
& \textbf{74.03}$\pm$\textbf{0.84} & \textbf{41.94}$\pm$\textbf{0.92} 
& \textbf{42.00}$\pm$\textbf{0.11} & \textbf{73.50}$\pm$\textbf{0.40}  \\

\bottomrule
\end{tabular}
}
\end{table*}

\begin{table*}
\setlength{\abovecaptionskip}{5pt}
\setlength{\belowcaptionskip}{-5pt}
\caption{Graph regression and classification results of different graph self-supervised learning on OGB datasets. Bold indicates
the best performance and underline means the runner-up. $\downarrow$ means lower the better and $\uparrow$ means higher the better.}

\scriptsize
\centering
\resizebox{16cm}{!}{
\begin{tabular}{C{1.5cm}C{1.3cm}C{1.3cm}C{1.3cm}C{1.3cm}C{1.3cm}C{1.3cm}C{1.3cm}}
\toprule
\textbf{Task} & \multicolumn{3}{c}{Regression (Metric: RMSE $\downarrow$)}    & \multicolumn{4}{c}{Classification (Metric: ROC-AUC\% $\uparrow$)}      \\
 \cmidrule(lr){2-4}   \cmidrule(lr){5-8}

\textbf{Dataset} & molesol & molipo & molfreesolv & molbace & molbbbp & molclintox & moltocx21  \\
\midrule

InfoGraph & 1.344$\pm$0.178 & 1.005$\pm$0.023 & 10.005$\pm$8.147 & 73.64$\pm$3.64 & 66.33$\pm$2.79 & 64.50$\pm$5.32 & 69.74$\pm$0.57 \\

GraphCL & 1.272$\pm$0.089 & 0.910$\pm$0.016 & 7.679$\pm$2.748 & 73.32$\pm$2.70 & 68.22$\pm$2.19 & 74.92$\pm$4.42 & 72.40$\pm$1.07 \\

MVGRL & 1.433$\pm$0.145 & 0.962$\pm$0.036 & 9.024$\pm$1.982 & 74.88$\pm$1.43 & 67.24$\pm$3.19 & 73.84$\pm$2.75 & 70.48$\pm$0.83 \\

JOAO & 1.285$\pm$0.121 & 0.865$\pm$0.032 & 5.131$\pm$0.782 & 74.43$\pm$1.94 & 67.62$\pm$1.29 & 71.28$\pm$4.12 & 71.38$\pm$0.92 \\

Sp$^2$GCL & 1.235$\pm$0.119 & 0.835$\pm$0.026 & 4.144$\pm$0.573 & 78.76$\pm$1.43 & \underline{68.72$\pm$1.53} & 80.88$\pm$3.86 & 73.06$\pm$0.75 \\
\midrule

GraphMAE & 1.050$\pm$0.034 & 0.850$\pm$0.022 & 2.740$\pm$0.233 & 79.14$\pm$1.31 & 66.55$\pm$1.78 & 80.56$\pm$5.55 & 73.84$\pm$0.58 \\

GraphMAE2 & 1.225$\pm$0.081 & 0.885$\pm$0.019 & 2.913$\pm$0.293 & 80.74$\pm$1.53 & 67.67$\pm$1.44 & 75.75$\pm$3.65 & 72.93$\pm$0.69 \\

StructMAE & 1.499$\pm$0.043 & 1.089$\pm$0.002 & 2.568$\pm$0.262 & 77.75$\pm$0.42 & 65.66$\pm$1.16 & 79.42$\pm$4.56 & 71.13$\pm$0.61 \\

AUG-MAE & 1.248$\pm$0.026 & 0.917$\pm$0.013 & 2.395$\pm$0.158 & 78.54$\pm$2.49 & 67.05$\pm$0.63 & 82.66$\pm$1.98 & 74.33$\pm$0.07 \\

GraphPAE & \underline{1.015$\pm$0.045} & \underline{0.810$\pm$0.018} & \underline{2.348$\pm$0.215} & \underline{81.11$\pm$1.24} & 68.56$\pm$0.71 & \underline{82.69$\pm$3.39} & \underline{74.46$\pm$0.54} \\
\midrule


\textbf{FC-GSSL(Ours)} & \textbf{1.001}$\pm$\textbf{0.042} & \textbf{0.789}$\pm$\textbf{0.014} 
& \textbf{2.279}$\pm$\textbf{0.233} & \textbf{82.69}$\pm$\textbf{0.81} 
& \textbf{68.92}$\pm$\textbf{1.73} & \textbf{83.40}$\pm$\textbf{1.67} & \textbf{74.61}$\pm$\textbf{0.80}  \\

\bottomrule
\end{tabular}
}
\end{table*}

\begin{table*}
\setlength{\abovecaptionskip}{5pt}
\setlength{\belowcaptionskip}{-5pt}
\caption{Quantum chemistry property results of transfer learning on QM9. The best and runner-up results are highlighted with
bold and underline, respectively.}

\scriptsize
\centering
\resizebox{16cm}{!}{
\begin{tabular}{C{1.5cm}C{0.6cm}C{0.6cm}C{0.6cm}C{0.6cm}C{0.6cm}C{0.6cm}C{0.6cm}C{0.6cm}C{0.6cm}C{0.6cm}C{0.6cm}C{0.6cm}C{0.6cm}C{0.6cm}}
\toprule
\textbf{Task} & $\mu$ & $\alpha$ & $\epsilon_{\text{homo}}$ & $\epsilon_{\text{lumo}}$ & $\Delta_{\epsilon}$ & $\textit{R}^2$ & ZPVE & $\textit{U}_0$ & $\textit{U}$ & $\textit{H}$ & $\textit{G}$ & $\textit{C}_v$ 
\\
\cmidrule(lr){2-13}

\textbf{Unit} & $\text{D}$ & $\textit{a}_0^3$ & $10^{-2}\text{meV}$ & $10^{-2}\text{meV}$ & $10^{-2}\text{meV}$ & $\textit{a}_0^2$ & $10^{-2}\text{meV}$  & $\text{meV}$ & $\text{meV}$ & $\text{meV}$ & $\text{meV}$ & $\text{cal/mol/K}$ \\
\midrule

GraphCL & 1.035 & 2.321 & 2.030 & 3.667 & 4.523 & 40.725 & 2.063 & 2.461 & 1.745 & 1.734 & 1.751 & 1.747 \\
GraphMAE & 1.030 & 2.924 & 2.407 & 6.373 & 4.813 & 41.955 & 4.623 & 1.411 & 2.207 & 2.208 & 2.207 & 2.200 \\
Mole-BERT & 1.031 & 1.918 & 1.477 & 4.127 & 4.240 & 44.374 & 2.190 & 2.532 & 2.509 & 2.511 & 2.516 & 2.508 \\
SimSGT & 1.064 & 2.413 & 2.837 & 4.227 & 4.107 & 40.504 & 2.127 & 1.948 & 2.420 & 2.416 & 2.416 & 2.410 \\
GraphPAE & \textbf{0.703} & \underline{0.879} & \underline{1.199} & \underline{2.141} & \underline{2.289} & \textbf{36.480} & \textbf{0.502} & \underline{0.510} & \underline{0.639} & \underline{0.639} & \underline{0.641} & \underline{0.643} \\

\midrule

\textbf{FC-GSSL(Ours)} & \underline{0.709} & \textbf{0.851} & \textbf{1.194} & \textbf{2.121} & \textbf{2.231} & \underline{36.659} & \underline{0.577} & \textbf{0.490} & \textbf{0.633} & \textbf{0.635} & \textbf{0.636} & \textbf{0.631} \\


\bottomrule
\end{tabular}
}
\end{table*}

\subsection{Node Classification}
\noindent\textbf{Settings.}
To validate the performance of the FC-GSSL model on the node classification task, we conducted experiments using accuracy as the evaluation metric on 6 representative heterogeneous graphs: BlogCatalog \cite{meng2019co}, Chameleon, Squirrel, Actor \cite{pei2020geom}, arXiv-year \cite{hu2020open}, and Penn94 \cite{traud2012social}. 
Of these, arXiv-year and Penn94 are large-scale graphs with node sizes exceeding 40,000, thus effectively assessing the scalability of the proposed method.

For a comprehensive evaluation, FC-GSSL is compared against a wide range of representative graph self-supervised learning methods. 
These methods are broadly divided into two categories:
\textbf{graph contrastive learning} methods, such as DGI \cite{velivckovic2018deep}, BGRL \cite{thakoor2021bootstrapped}, MVGRL \cite{hassani2020contrastive}, CCA-SSG \cite{zhang2021canonical}, and Sp$^2$GCL \cite{bo2023graph}; and \textbf{graph autoencoders}, including VGAE \cite{kipf2016variational}, GraphMAE \cite{hou2022graphmae}, GraphMAE2 \cite{hou2023graphmae2}, MaskGAE \cite{li2023s}, S2GAE \cite{tan2023s2gae}, AUG-MAE \cite{wang2024rethinking}, and GraphPAE \cite{liu2025graph}.
Of these, Sp$^2$GCL and GraphPAE extract high-frequency signals in graphs through graph spectral filtering and positional encodings, respectively. 
Using them as baselines can better demonstrate the effectiveness of our model in integrating high-frequency signals.

For the node classification task,
we employ a GAT encoder with 4 attention heads, 1024 hidden units per layer, and the number of layers selected from $\{2, 3\}$. 
Node and edge features are decoded using two-layer MLPs with ReLU activations. 
For the downstream node classification task, the encoder weights are frozen, and a linear classifier is trained on the generated node embeddings. 
All models are optimized with Adam, and we report the average results over 5 experimental runs per graph.

\noindent\textbf{Results.}
Table 1 shows the experimental results. From it we observe that: (1) Models Sp$^2$GCL and GraphPAE, which focus on high-frequency signals, have a clear advantage over other baseline models in the same category on most datasets; 
(2) FC-GSSL outperforms all baselines, achieving the best overall performance.


The reasons for these results are:
(1) Models emphasizing high-frequency signals better capture local structure, enriching node representations and improving classification.
(2) FC-GSSL enhances existing methods by addressing the underutilization of high-frequency signals and over-reliance on specific components. It uses high-frequency-biased corrupted graphs as autoencoder inputs, with low-frequency and general features as targets, forcing effective multi-band signal integration. Generating diverse corrupted graphs and aligning their representations reduces sensitivity to specific components, improving generalization. Consequently, FC-GSSL strengthens local feature extraction and achieves superior node classification performance.

\subsection{Graph Prediction}
\noindent\textbf{Settings.}
The evaluation of FC-GSSL for graph-level tasks is conducted across 7 OGB datasets \cite{hu2020open}. 
This benchmark includes 3 graph regression and 4 graph classification problems. 
In all experiments, the standard public splits are employed to guarantee a fair comparison. Model performance is measured by MSE for regression and ROC-AUC for classification. 

Our evaluation on graph prediction tasks involved a comparative analysis with benchmark methods from two categories: \textbf{graph contrastive learning} methods, such as InfoGraph \cite{sun2019infograph}, GraphCL \cite{you2020graph}, MVGRL \cite{hassani2020contrastive}, JOAO \cite{you2021graph}, and Sp$^2$GCL \cite{bo2023graph}; and \textbf{graph autoencoders}, including GraphMAE \cite{hou2022graphmae}, GraphMAE2 \cite{hou2023graphmae2}, StructMAE \cite{liu2024mask}, AUG-MAE \cite{wang2024rethinking}, and GraphPAE \cite{liu2025graph}.

For the graph prediction tasks, we employ a two-layer GatedGCN encoder (hidden dimension $d_h$=300) and two-layer MLP decoders. 
During evaluation, node representations from the frozen encoder are pooled to form graph representations, which are then fed into a linear classifier. 
The model is trained using the Adam optimizer, and the reported results represent the average and standard deviation over 5 experimental runs per graph.

\noindent\textbf{Results.}
Table 2 presents the performance of the FC-GSSL model on graph prediction tasks. 
From the results, it can be observed that the model achieves excellent performance on both graph regression and graph classification tasks, significantly outperforming existing baseline methods based on contrastive learning and graph autoencoders.

The reasons for these results are that graph prediction tasks typically involve structures of relatively small scale, which require more fine-grained extraction of local features. 
Compared to other baseline models, FC-GSSL places greater emphasis on utilizing effective high-frequency signals while filtering out ineffective ones, allowing it to capture local features in graphs more precisely and thereby improving the accuracy of graph prediction tasks.

\subsection{Transfer Learning}
\noindent\textbf{Settings.}
We evaluated FC-GSSL's generalization via transfer learning for molecular property prediction. 
The encoder was pre-trained on 2 million ZINC15 \cite{sterling2015zinc} molecules and fine-tuned on QM9 \cite{wu2018moleculenet} to predict quantum properties. 
We compare it with GraphCL \cite{you2020graph}, GraphMAE \cite{hou2022graphmae}, Mole-BERT \cite{xia2023mole}, SimSGT \cite{liu2023rethinking}, and GraphPAE \cite{liu2025graph}. 
The encoder uses a five-layer GatedGCN (hidden dim 300), followed by a two-layer MLP for prediction. 
Fine-tuning trains both the encoder and the MLP on labeled downstream task data.
QM9 is split 8:1:1 into training/validation/test sets. 
Experiments use Adam optimizer and are repeated 5 times for average results.

\noindent\textbf{Results.}
Table 3 shows the transfer learning results. The FC-GSSL model outperforms baseline methods on most tasks. 
This is because it focuses not on extracting high-frequency signals, but on effectively fusing signals of different frequencies. 
We also use contrastive learning to reduce the model’s dependence on specific frequencies, improving its generalization. As a result, FC-GSSL can learn high-quality frequency patterns and achieves more reliable and transferable performance than other transfer learning methods.

\begin{table}
\setlength{\abovecaptionskip}{5pt}
\setlength{\belowcaptionskip}{-5pt}
\caption{Ablation studies on key components of FC-GSSL.}

\scriptsize
\centering
\resizebox{8.5cm}{!}{
\begin{tabular}{C{1cm}C{0.7cm}C{0.7cm}C{0.7cm}C{0.7cm}C{0.7cm}C{0.7cm}}
\toprule
\multirow{2}*{\textbf{Dataset}}  & \multicolumn{2}{c}{Node Classification ($\uparrow$)}  & \multicolumn{2}{c}{Graph
Regression ($\downarrow$)}  & \multicolumn{2}{c}{Graph Classification ($\uparrow$)}      \\
\cmidrule(lr){2-3} \cmidrule(lr){4-5}   \cmidrule(lr){6-7}

& Squirrel  & Actor & molesol  & molipo &  molbace  & molbbbp  \\
\midrule

\textbf{w/o cn} & 73.24 & 40.88 & 1.022 & 0.798 & 79.35 & 67.72 \\
\textbf{w/o ce} & 72.48 & 41.52 & 1.025 & 0.805 & 80.09 & 67.01 \\
\textbf{w/o cne} & 71.93 & 40.58 & 1.028 & 0.816 & 78.80 & 66.50 \\
\textbf{w/o so} & 73.39 & 41.35 & 1.030 & 0.806 & 78.54 & 66.80 \\
\textbf{w/o sa} & 73.21 & 38.43 & 1.016 & 0.804 & 79.97 & 67.35 \\
\textbf{w/o soa} & 72.34 & 37.58 & 1.054 & 0.808 & 77.65 & 66.20 \\

\midrule

\textbf{FC-GSSL} & \textbf{74.03} & \textbf{41.94}
& \textbf{1.001} 
& \textbf{0.789}
& \textbf{82.69} 
& \textbf{68.92}  \\

\bottomrule
\end{tabular}
}
\end{table}

\subsection{Ablation Studies}
In this section, we validate the effectiveness of our design for generating and aligning corrupted graphs based on frequency contribution by employing variant experiments evaluated on 3 tasks across 6 datasets.
(1) \textbf{w/o cn}: Eliminate corrupted nodes based on low-frequency signal contributions and adopt a random corruption strategy.
(2) \textbf{w/o ce}: Eliminate corrupted edges based on low-frequency signal contributions and adopt a random corruption strategy.
(3) \textbf{w/o cne}: Eliminate all operations corrupted by low-frequency signal contributions and adopt a random corruption strategy.
(4) \textbf{w/o so}: 
Eliminate intersection and union operations across sampling strategies, using the value-based corrupted graph as input and the rank-based corrupted graph as the contrastive view.
(5) \textbf{w/o sa}: Eliminate alignment between representations from different corrupted graphs.
(6) \textbf{w/o soa}: Eliminate intersection and union operations across sampling strategies, and eliminate alignment between representations from different corrupted graphs.









Based on the ablation study results as shown in Table 4, we can draw the following conclusions:
(1) Comparing FC-GSSL with its variants \textbf{w/o cn}, \textbf{w/o ce}, and \textbf{w/o coe} shows that using high-frequency corrupted graphs as autoencoder inputs, guided by low-frequency and general features, effectively directs the model's frequency learning process and significantly improves high-frequency utilization.
(2) Comparing FC-GSSL with its variants \textbf{w/o so}, \textbf{w/o sa}, and \textbf{w/o soa} shows that generating diverse corrupted graphs and aligning their node representations can suppress irrelevant frequency dependencies and reduce interference from unrelated signals, thereby enhancing model generalization.





\begin{figure}[t]
\setlength{\abovecaptionskip}{0pt}
\setlength{\belowcaptionskip}{-5pt}
\centering
\includegraphics[scale=0.34]{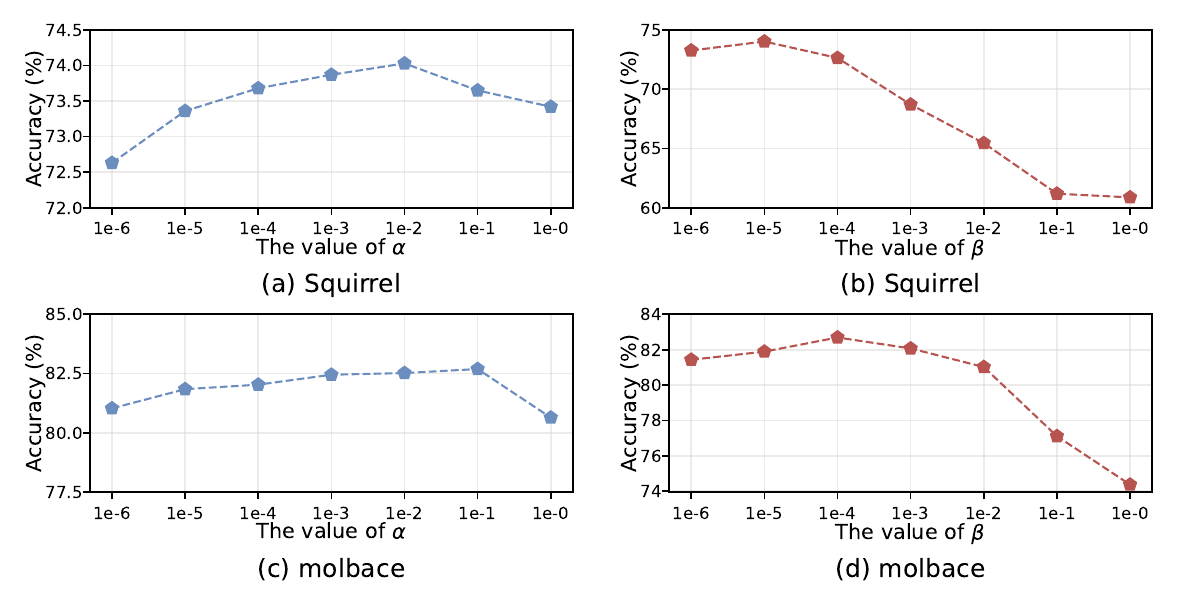}
\caption{The Influence of the loss weights $\alpha$ and $\beta$.}
\label{fig:label}
\end{figure}

\begin{figure}[t]
\setlength{\abovecaptionskip}{0pt}
\setlength{\belowcaptionskip}{-5pt}
\centering
\includegraphics[scale=0.34]{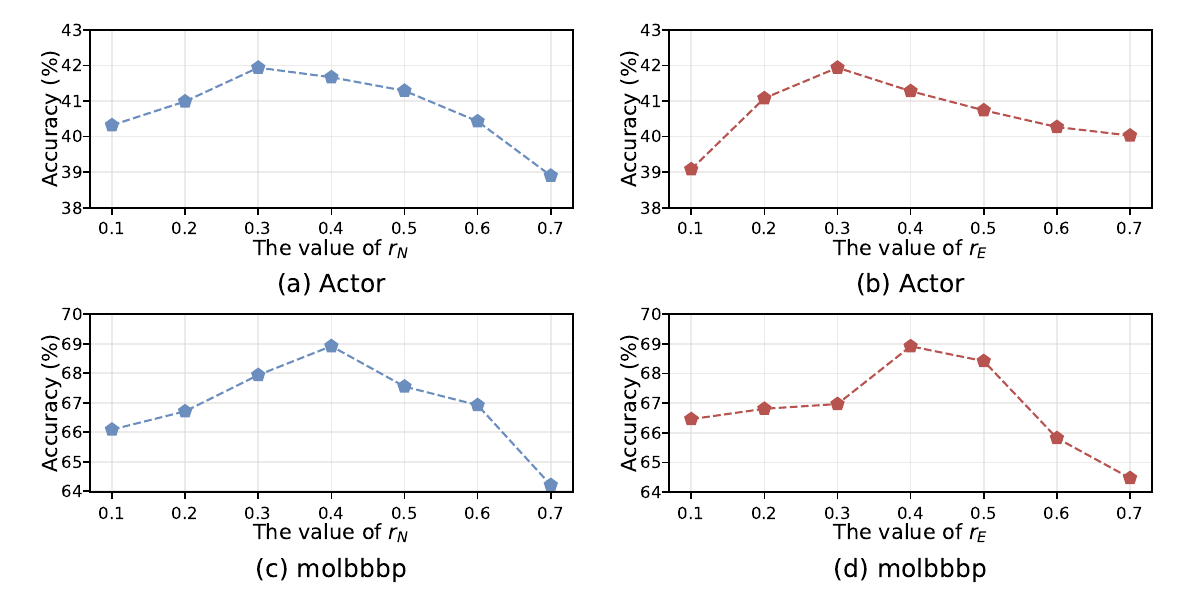}
\caption{The Influence of the sampling rates $r_N$ and $r_E$.}
\label{fig:label}
\end{figure}

\subsection{Hyperparameter Analysis}

We perform a thorough parameter analysis on the loss weights $\alpha$ and $\beta$, the sampling rates $r_N$ and $r_E$. 
The experimental results are shown in Fig. 2 and Fig. 3.

The experimental results lead to the following conclusions:
(1) As shown in Fig. 2, the change in loss weight $\alpha$ has less impact on performance than weight $\beta$. This is because the tasks of reconstructing the edge structure and node features are similar; consequently, the additional information provided by the $\mathcal{L}_{edge}$ term, which is controlled by $\alpha$, is limited. 
Furthermore, edge corruption helps align representations from different corrupted graphs, enhancing the model's sensitivity to the $\mathcal{L}_{align}$ term, which is controlled by $\beta$.
(2) As shown in Fig. 3, the model is more sensitive to sampling rate changes on the molbbbp dataset than on the Actor dataset. This is because the limited scale of the molbbbp dataset restricts the model's ability to learn from deeper neighbors, thereby making the sampling rate more critical.

The above experiments demonstrate that FC-GSSL is an effective graph self-supervised learning model. We conduct extensive experiments on 14 datasets across tasks including node classification, graph prediction, and transfer learning to validate the efficacy of our approach. Ablation studies further confirm that each component of the model contributes positively to performance improvement. Finally, we analyze key hyperparameters to provide deeper insights into their impact on the model’s performance.

\section{Conclusion}
In this paper, we have proposed the \textbf{F}requency-\textbf{C}orrupt Based \textbf{G}raph \textbf{S}elf-\textbf{S}upervised \textbf{L}earning (FC-GSSL) algorithm. 
Our approach constructed high-frequency-biased corrupted graphs as inputs to graph autoencoders, enhancing high-frequency information utilization. Additionally, we designed multiple sampling strategies to generate diverse corrupted graphs and aligned their node representations, preventing overreliance on specific high-frequency patterns and improving generalization capability. 
Experimental results from multiple tasks on 14 graph datasets have demonstrated the effectiveness of our proposed approach.

\section{ACKNOWLEDGMENTS}
This work is sponsored by 
the National Natural Science Foundation of China (Nos. 62506192, 62202253,
62272254, 62172249, 61973180), 
and the Natural Science Foundation of Shandong Province (Nos. ZR2024QF199, ZR2022MF326, ZR2021MF092,
ZR2021QF074).


\bibliographystyle{ACM-Reference-Format}
\bibliography{ref.bib}

\appendix

\section{Model Details}

\subsection{Training Procedure of FC-GSSL}
\label{fcgssl}

In this section, we will illustrate the training process of the FC-GSSL in detail.
The pseudo-code of FC-GSSL is shown in
Algorithm~\ref{algfc}. 

\begin{itemize}[leftmargin=10pt]
\item \textbf{Step 1.} 
We preprocess the graph, compute the top-$K$ eigenvectors $\textbf{U}$ and eigenvalues $\Lambda$, and then utilize them to calculate the low-frequency signal contributions $C_{N}$ and $C_{E}$, as well as the edge features $\textbf{E}$, followed by the initialization of the network model (\textit{lines 1-2}).

\item \textbf{Step 2.} 
Within each training epoch, we generate multiple corrupted graphs using different sampling strategies, which are then fed into the encoder to generate and align their node representations
(\textit{lines 4-8}).



\item \textbf{Step 3.}
The generated node representations are fed into the decoders to formulate reconstruction losses (lines 9-12), with the overall loss function being constructed and minimized to update the model parameters (lines 13-14).

\item \textbf{Step 4.}
Return the trained encoder (\textit{line 15}).

\end{itemize}

\begin{algorithm}[h]
    \caption{FC-GSSL}
    \LinesNumbered 
    \label{algfc}
    \footnotesize

    \KwIn{
    Graph $\mathcal{G}=\{ \mathcal{V}, \mathcal{E}, \textbf{X} \}$,
    sampling rates $r_N$ and $r_E$,
    and epochs $\psi$.
    }
    
    \textbf{Preprocess:} Compute the top-$K$ eigenvectors $\textbf{U}$ and eigenvalues $\Lambda$, and then utilize them to calculate the low-frequency signal contributions $C_{N}$ and $C_{E}$, as well as the edge features $\textbf{E}$.

    \textbf{Init:} Encoder $f_{enc}$, decoder $f_{dec}^N$, decoder $f_{dec}^E$, and learnable token $\textbf{x}_{[M]}$.

    \For{$i \leftarrow 1$ to $\psi$}{
        $\mathcal{P}_N,\mathcal{Q}_N,\mathcal{P}_E,\mathcal{Q}_E$
        $\leftarrow$ Generate sampling sets based on contribution values $C_{N}$ and $C_{E}$, with sampling rates $r_N$ and $r_E$ \;
        
        $\mathcal{S}_N,\mathcal{S}_E,\mathcal{S}_C$ $\leftarrow$ Generate the corrupted item sets from sampling sets $\mathcal{P}_N,\mathcal{Q}_N,\mathcal{P}_E,$ and $\mathcal{Q}_E$ \;
        
        $\mathcal{G}_N,\mathcal{G}_E,\mathcal{G}_C$ $\leftarrow$ Generate the corrupted corruption graph from the corrupted item sets $\mathcal{S}_N,\mathcal{S}_E,$ and $\mathcal{S}_C$ \;

        $\textbf{X}^{(L)}_N$, $\textbf{X}^{(L)}_E$, $\textbf{X}^{(L)}_C$ 
        $\leftarrow$ Using $\mathcal{G}_N,\mathcal{G}_E,$ and $\mathcal{G}_C$ as input to the encoder $f_{enc}$ to generate the corresponding node representations \;

        $\mathcal{L}_{align}$  $\leftarrow$ Generate an alignment loss by aligning node representations $\textbf{X}^{(L)}_N$, $\textbf{X}^{(L)}_E$, and $\textbf{X}^{(L)}_C$ \;

        $\hat{\textbf{X}}$ $\leftarrow$  Use $\textbf{X}^{(L)}_N$ as the input of the decoder $f_{dec}^N$ to reconstruct node features \;

        $\mathcal{L}_{node}$  $\leftarrow$ Generate the loss for node feature reconstruction based on $\hat{\textbf{X}}$ and $\textbf{X}$ \;

        $\hat{\textbf{E}}$ $\leftarrow$  Use $\textbf{E}^{(L)}_N$ as the input of the decoder $f_{dec}^E$ to reconstruct edge features \;

        $\mathcal{L}_{edge}$  $\leftarrow$ Generate the loss for edge structure reconstruction based on $\hat{\textbf{E}}$ and $\textbf{E}$ \;

        $\mathcal{L}$  $\leftarrow$ Generate the overall loss from $\mathcal{L}_{node}$, $\mathcal{L}_{edge}$, and $\mathcal{L}_{align}$ \;

        Update $f_{enc}$, $f_{dec}^N$, and $f_{dec}^E$ by minimizing $\mathcal{L}$ \;
        
    }

    \Return Trained encoder $f_{enc}$ \;

\end{algorithm}

\subsection{Time Complexity Analysis}
\label{tca}

We analyze the time complexity of various components in FC-GSSL from multiple aspects:
(1) To improve computational efficiency during eigen-decomposition, we consider only the smallest $K$ eigenvalues, reducing the time complexity to $\mathcal{O}(N^2K)$.
(2) The time complexity in the frequency contribution analysis stage is $\mathcal{O}( \lvert \mathcal{E}  \rvert K^2 )$, where $\lvert \mathcal{E}  \rvert$ represents the number of edges.
(3) The time complexity of the encoder is $\mathcal{O}( \lvert \mathcal{E}  \rvert d )$.
(4) Since the decoder is implemented solely using MLP, its time complexity is approximately $\mathcal{O}( d^2 )$.

\section{Experimental Details}

\begin{table}
\setlength{\abovecaptionskip}{2.5pt}
\setlength{\belowcaptionskip}{-5pt}
\caption{Dataset statistics for node classification.}

\scriptsize
\centering
\resizebox{8.5cm}{!}{
\begin{tabular}{L{1cm}C{0.55cm}C{0.55cm}C{0.55cm}C{0.55cm}L{1.4cm}}
\toprule

\textbf{Dataset} & $|\mathcal{V}|$  & $|\mathcal{E}|$ & $|\mathcal{F}|$  & $|\mathcal{C}|$ &  {Train/Valid/Test}   \\
\midrule

BlogCatalog &  5,196 & 343,486 & 8,189 & 6 & 120/1,000/1,000  \\
Chameleon   &  2,277 & 62,792 & 2,325&  5 & 1,092/729/456  \\
Squirrel &  5,201 &  396,846 &  2,089 &  5 &  2,496/1,664/1,041  \\
Actor &  7,600 &  53,411 &  932 &  5 &  3,648/2,432/1,520  \\
arXiv-year &  169,343 &  2,315,598 &  128 &  5 &  84671/42335/42337  \\
Penn94 &  41,554 &  2,724,458 &  4,814 &  2 &  19,407/9,703/9,705  \\

\bottomrule
\end{tabular}
}
\end{table}

\begin{table}
\setlength{\abovecaptionskip}{2.5pt}
\setlength{\belowcaptionskip}{-5pt}
\caption{Dataset statistics for graph prediction and transfer learning.}

\scriptsize
\centering
\resizebox{8.5cm}{!}{
\begin{tabular}{L{0.8cm}C{0.55cm}C{0.7cm}C{0.7cm}C{1cm}L{1.2cm}L{1cm}}
\toprule

\textbf{Dataset} & $|\mathcal{G}|$  & {Avg. $|\mathcal{V}|$} & {Avg. $|\mathcal{E}|$}  & $|\mathcal{C}|$ & {Task} &  {Metric}   \\
\midrule

molesol & 1,128 & 13.3 & 13.7 & 1 & Regression & RMSE \\
mollipo & 4,200 & 27.0 & 29.5 & 1 & Regression & RMSE \\
molfreesolv & 642 & 8.7 & 8.4 & 1 & Regression & RMSE \\
molbace& 1,513 & 34.1 & 36.9 & 1 & Binary Class. & ROC-AUC \\
molbbbp & 2,039 & 24.1 & 26.0 & 1 & Binary Class. & ROC-AUC \\
molcintox & 1,477 & 26.2 & 27.9 & 2 & Binary Class. & ROC-AUC \\
moltox21 & 7,831 & 18.6 & 19.3 & 12 & Binary Class. & ROC-AUC \\

\midrule

ZINC15 & 2,000,000 & 26.62 & 57.72 &  - & Pre-Training &  - \\
QM9 & 1,177,631 &  8.80  &  18.81 &  12 Targets &  Finetuning &  MAE \\

\bottomrule
\end{tabular}
}
\end{table}

\subsection{Statistics of Datasets}
\label{esd}

We benchmarked FC-GSSL on tasks including node classification, graph prediction, and transfer learning. Experiments were conducted on 6 node classification and 7 graph-level prediction datasets. For transfer learning, we employed ZINC15 and QM9, splitting the latter into training, validation, and test sets with an $80\%/10\%/10\%$ ratio according to \cite{liu2023rethinking}. Public data splits were adopted for the other datasets.
Table 5 presents the statistics of the node classification datasets, with
$|\mathcal{V}|$, $|\mathcal{E}|$, $|\mathcal{F}|$, and $|\mathcal{C}|$ denoting the number of nodes, edges, feature dimension, and classes, respectively.
Table 6 presents the statistics of the datasets used for graph prediction and transfer learning, where $|\mathcal{G}|$ represents the number of graphs.

\begin{table}
\setlength{\abovecaptionskip}{2.5pt}
\setlength{\belowcaptionskip}{-5pt}
\caption{Hyperparameters of node classification.}

\scriptsize
\centering
\resizebox{8.5cm}{!}{
\begin{tabular}{L{1cm}C{0.55cm}C{0.55cm}C{0.55cm}C{0.55cm}C{0.55cm}C{0.55cm}C{0.55cm}}
\toprule

\textbf{Dataset} & lr & $\alpha$  & $\beta$ & $r_N$  & $r_E$ & $K$ &  $K_{e}$   \\
\midrule

BlogCatalog   &  0.001    &  0.01 & 0.00001   & 0.3 & 0.3        & $|\mathcal{V}|$ & 50  \\

Chameleon   &  0.001    &  0.01 & 0.0001   & 0.3 & 0.3        & $|\mathcal{V}|$ & 50  \\
Squirrel    &  0.001    &  0.01 & 0.00001  & 0.3 & 0.3  & $|\mathcal{V}|$ & 50 \\
Actor       &  0.0005   &  0.01 & 0.0001   & 0.3 & 0.3   & $|\mathcal{V}|$ & 50 \\
arXiv-year  &  0.001    &  1 & 0.01        &  0.1 & 0.1       &  1000  & 100 \\
Penn94      &  0.0025   &  1 & 0.1         &  0.25 & 0.25        &  1000  & 200 \\

\bottomrule
\end{tabular}
}
\end{table}

\begin{table}
\setlength{\abovecaptionskip}{2.5pt}
\setlength{\belowcaptionskip}{-5pt}
\caption{Hyperparameters of graph prediction.}

\scriptsize
\centering
\resizebox{8.5cm}{!}{
\begin{tabular}{L{1cm}C{0.55cm}C{0.55cm}C{0.55cm}C{0.55cm}C{0.55cm}C{0.55cm}C{0.55cm}C{0.55cm}}
\toprule

\textbf{Dataset} & pooling & lr & $\alpha$  & $\beta$ & $r_N$  & $r_E$ & $K$ &  $K_{e}$   \\
\midrule

molesol & sum & 0.0005          & 0.01 & 0.00001    & 0.75 & 0.75 & 8  &  8 \\
mollipo & sum & 0.0005          & 0.001 & 0.0001    & 0.25 & 0.25 & 30  &  30  \\
molfreesolv & sum  & 0.00002    & 0.01 & 0.0001     & 0.5  & 0.25 & 15  &  15 \\
molbace & mean & 0.001          & 0.1 & 0.0001      & 0.3  & 0.3  & 30  &  30 \\
molbbbp & mean & 0.001          & 0.01 & 0.00001    & 0.4  & 0.4  & 30  &  30 \\
molcintox & mean & 0.001        & 0.1 & 0.001       & 0.3  & 0.3  & 30  &  30 \\
moltox21 & mean & 0.0001        & 0.2 & 0.00001     & 0.25 & 0.25 & 8  &  8 \\

\bottomrule
\end{tabular}
}
\end{table}

\subsection{Hyperparameters}
\label{ehp}
The key hyperparameters for node classification and graph prediction are provided in Table 7 and Table 8, respectively.
Specifically, parameter $K$ controls the number of low-frequency signals considered for computing contributions, whereas $K_e$ controls the number used in the generation of positional encodings and edge features.
In transfer learning, we pre-trained FC-GSSL for 100 epochs on 2 million molecules sampled from the ZINC15 dataset. 
During pretraining, 
$\alpha$ was set to 0.01, 
$\beta$ to 0.0001, 
$r_N$ and 
$r_E$ both to 0.35, and 
$K$ and 
$K_{e}$ both to 6.
In the fine-tuning phase, the encoder and its additional MLP were jointly trained on the QM9 dataset, with an initial learning rate of 0.001.
For all experiments, the hyperparameter 
$\gamma$ is searched in the range of 
$\{ 1.0, 2.0, 3.0\}$, and the hyperparameter 
$\tau$ is set to 0.2.
More detailed hyperparameter configurations are available in the source code at  
https://github.com/rookitkitlee/FC-GSSL.

\end{document}